# Unsupervised Word Mapping Using Structural Similarities in Monolingual Embeddings

**Hanan Aldarmaki**  **Mahesh Mohan**  **Mona Diab**

Department of Computer Science
The George Washington University
`{aldarmaki;mahesh_mohan;mtdiab}@gwu.edu`

## Abstract

Most existing methods for automatic bilingual dictionary induction rely on prior alignments between the source and target languages, such as parallel corpora or seed dictionaries. For many language pairs, such supervised alignments are not readily available. We propose an unsupervised approach for learning a bilingual dictionary for a pair of languages given their independently-learned monolingual word embeddings. The proposed method exploits local and global structures in monolingual vector spaces to align them such that similar words are mapped to each other. We show empirically that the performance of bilingual correspondents learned using our proposed unsupervised method is comparable to that of using supervised bilingual correspondents from a seed dictionary.

## 1 Introduction

The working hypothesis in distributional semantics is that the meaning of a word can be inferred by its distribution, or co-occurrence, around other words. The validity of this hypothesis is most evident in the performance of distributed vector representations of words, i.e word embeddings, that are automatically induced from large text corpora (Bengio et al., 2003; Mikolov et al., 2013b). The qualitative nature of these embeddings can be demonstrated through empirical evidence of regularities that reflect certain semantic relationships. Words in the vector space are generally clustered by meaning, and the distances between words and clusters reflect semantic or syntactic relationships, which makes it possible to perform arithmetic on word vectors for analogical reasoning and semantic composition (Mikolov et al., 2013b). For example, in a vector space $V$ where $f = V(``france")$, $p = V(``paris")$, and $g = V(``germany")$, the distance $f - p$ reflects the country-capital relationship, and $g + f - p$ results in a vector closest to $V(``berlin")$. Named entities and inflectional morphemes are particularly amenable to vector arithmetic, while derivational morphology, polysemy, and other nuanced semantic categories result in lower performance in analogy questions (Finley et al., 2017).

The extent of these semantic and syntactic regularities is difficult to assess intrinsically, and the performance in analogical reasoning can be partially attributed to the clustering of the words in question (Linzen, 2016). If meaning is encoded in the relative distances among word vectors, then the structure within vector spaces should be consistent across different languages given that the datasets used to build them express similar content. In (Rapp, 1995), a simulation study showed that similarity in word co-occurrence patterns within unrelated German and English texts is correlated with the number of corresponding word positions in the monolingual co-occurrence matrices. More recently, (Mikolov et al., 2013a) showed that a linear projection can be learned to transform word embeddings from one language into the vector space of another using a medium-size seed dictionary, which demonstrates that the multilingual vector spaces are at least related by a linear transform. This makes it possible to align word embeddings of different languages in order to be directly comparable within the same seman-tic space. Such cross-lingual word embeddings can be used to expand dictionaries or to learn language-independent classifiers.

A number of methods have been proposed recently for learning cross-lingual word embeddings with various degrees of supervision, ranging from word-level alignment using bilingual dictionaries (Ammar et al., 2016), sentence-level alignment using parallel corpora (Gouws et al., 2015; Klementiev et al., 2012), or document alignment using cross-lingual topic models (Vulić and Moens, 2015; Vulić and Moens, 2012). Using such alignments, especially large parallel corpora or sizable dictionaries, high-quality bilingual embeddings can be obtained (Upadhyay et al., 2016). In addition, a number of methods have been proposed for expanding dictionaries using a small initial dictionary with as few as a hundred entries (Haghighi et al., 2008).

However, such alignments are not available for all languages and dialects, and while a small dictionary might be feasible to acquire, discovering word mappings with no prior knowledge whatsoever is valuable. Intuitively, if the monolingual corpora express similar aspects of the world, there should be enough structure within the vector space of each language to recover the mappings in a completely unsupervised manner. In this paper, we propose a novel approach for learning a transformation between monolingual word embeddings without the use of prior alignments. We show empirically that we can recover mappings with high accuracy in two language pairs: a close language pair, French-English; and a distant language pair, Arabic-English. The proposed method relies on the consistent regularities within monolingual vector spaces of different languages. We extract initial mappings using spectral embeddings that encode the local geometry around each word, and we use these tentative pairs to seed a greedy algorithm which minimizes the differences in global pair-wise distances among word vectors. The retrieved mappings are then used to fit a linear projection matrix to transform word embeddings from the source to the target language.

## 1.1 Related Work

Few models have been proposed for extracting dictionaries or learning bilingual embeddings without the use of any prior alignment. For languages that share orthographic similarities, lexical features such as the normalized edit distance between source and target words can be used to extract a seed lexicon for bootstrapping the bilingual dictionary induction process (Hauer et al., 2017). In (Diab and Finch, 2000), unsupervised mappings were extracted by preserving pairwise distances between word co-occurrence representations from two comparable corpora. The model was only evaluated mono-lingually, where two sections of a corpus were used for collecting co-occurrence statistics separately, and an iterative training algorithm was then used to retrieve the mapping of English words to themselves. Only punctuation marks were used to seed the learning and high accuracy results were reported. However, the method was not evaluated cross-lingually. We observed experimentally that punctuation marks—and function words in general—are insufficient to map words cross-lingually since they have different distributional profiles in different languages due to their predominant syntactic role.

Another unsupervised approach has been recently proposed using adversarial autoencoders (Barone, 2016) where a transformation is learned without a seed by matching the distribution of the source word embeddings with the target distribution. Preliminary investigation showed some correct mappings but the results were not comparable to supervised methods. Recents efforts using carefully-tuned adversarial methods report encouraging results comparable to supervised methods (Zhang et al., 2017; Conneau et al., 2017). In (Kiela et al., 2015), bilingual lexicon induction is achieved by matching visual features extracted from images that correspond to each word using a convolutional neural network. The image-based approach performs particularly well for words that express concrete rather than abstract concepts, and provides a convenient alternative to linguistic supervision when corresponding images are available.

The unsupervised mapping problem arises in other contexts where an optimal alignment between two isomorphic point sets is sought. In image registration and shape recognition, various efficient methods can be used to find an optimal alignment between two sets of low-dimensional points that correspond to images with various degrees of deformation (Myronenko and Song, 2010; Chi et al., 2008). In manifold learning, two sets of related high-dimensional points are projected into a shared lower dimensional space where the points can be compared and mapped to one other, such as the alignment of isomorphic protein structures (Wang and Mahadevan, 2009) and cross-lingual document alignment with unsupervised topic models (Diaz and Metzler, 2007; Wang and Mahadevan, 2008).

## 2 Background

### 2.1 Skip-gram Word Embeddings with Subword Features

In the skip-gram model presented in (Mikolov et al., 2013b), a feed-forward neural network is trained to maximize the probability of all words within a fixed window around a given word. Formally, given a word $w$ in a vocabulary $W$, the objective of the skip-gram model is to maximize the following log-likelihood:

$$\sum_{c \in C_w} \log p(c|w)$$

where $C_w$ is the set of words in the context of $w$. The words are represented as one-hot vectors of size $|W|$ that are projected into dense vectors of size $d$. Over a large corpus, the $d$-dimensional word projections encode semantic and syntactic features that are not only useful for maximizing the above probability, but also serve as general-purpose representations for words.

In (Bojanowski et al., 2016), a word vector is represented as the sum of its character n-grams which helps account for inflectional variations within a language, especially for morphologically rich languages where less frequent inflections are less likely to have good representations using only word-level features. Using n-grams helps account for lexical similarities among words within the same language; independently-learned embeddings with no explicit alignment would still have unrelated n-gram representations even if the languages share lexical similarities. We will refer to this model as the subword skip-gram.

### 2.2 Linear Transformation of Word Embeddings

Given word embeddings in two languages $X$ and $Y$, and a dictionary of $(source, target)$ word pairs with embeddings $x_s$ and $y_t$, respectively, a transformation matrix $T$, such that $y_t = Tx_s$, can be estimated with various degrees of accuracy (Mikolov et al., 2013a). Large, accurate dictionaries result in better transformations, but a good fit can also be obtained using a few thousand word pairs even in the presence of noise (see Section 4.3.4 for an empirical demonstration).

Formally, given a dictionary of $n$ word pairs, $(x_i, M(x_i))$, where $i = 1, ..., n$, and $M$ is a mapping from $X$ to $Y$, the linear transformation matrix $\hat{T}$ is learned by minimizing the following cost function

$$\hat{T} = \arg\min_T \sum_{i=1}^{n} \parallel Tx_i - M(x_i) \parallel^2 \quad (1)$$

After learning $\hat{T}$, the translation of new source words can be retrieved by transforming the word vector first, then finding its nearest neighbor in the target vocabulary.

## 3 Unsupervised Word Mapping

Learning an accurate transformation between word embeddings as described in Section 2.2 requires a seed dictionary of reasonable size. We propose a method that bypasses this requirement by learning to align the monolingual embeddings in an unsupervised manner. The underlying assumption is that word embeddings across different languages share similar local and global structures that characterize language-independent semantic features. For example, the distance between the words *monday* and *week* in English should be relatively similar to the distance between *lundi* and *semaine* in French. We attempt to recover the correspondences between different languages by exploiting these structural similarities.

Our approach consists of two main steps. In the first step (Section 3.1), we extract initial mappings using spectral features that encode the geometry of the local neighborhood around a point in the vector space. In the second step (Section 3.2), we it-eratively refine the correspondences using a greedy optimization algorithm, which we refer to as Iterative Mapping (IM for short). IM is a variation on the word mapping model in (Diab and Finch, 2000). The model does not make language-specific assumptions, making it suitable for learning crosslingual correspondences. We then use these correspondences to learn a linear transformation between the source and target embeddings, as described in Section 2.2.

### 3.1 Estimating Initial Correspondences

To analyze local structures in monolingual vector spaces, we treat each word embedding as a point in a high-dimensional space and further embed each point into a local invariant feature space, as proposed in (Chi et al., 2008) for affine registration of image point sets. The local invariant features are produced through eigendecomposition of the k-nearest-neighbor (*knn*) graph for each point in the vector space as described below.

For a word embedding $w$, we construct its *knn* adjacency graph, $A_w$, such that $A_w$ is a $k \times k$ matrix that contains the pair-wise similarities among $w$'s $k$-nearest neighbors, including $w$ itself. To embed the adjacency matrix in a permutation-invariant space, $w$ is mapped to a feature vector $v_w$ that contains the sorted eigenvalues of $\mathcal{L}_w$, which is defined as,

$$\mathcal{L}_w = \mathbf{I}_k - \begin{bmatrix} f(d_{11}) & \dots & f(d_{1k}) \\ \vdots & \vdots & \vdots \\ f(d_{k1}) & \dots & f(d_{kk}) \end{bmatrix}$$

where $f(d_{ij}) = \exp(-d_{ij}^2/2\sigma^2)$ is the Gaussian similarity function and $d_{ij}$ is the Euclidean distance between points $i$ and $j$. We will refer to the vectors of sorted eigenvalues as spectral embeddings.

After extracting these local features for all points in $X$ and $Y$, each point $p$ in $X$ is mapped to its nearest neighbor $q$ in $Y$ using the Euclidean distance between their spectral embeddings. To minimize the spurious effect of hubs—points that tend to be nearest neighbors to a large number of other points (Radovanović et al., 2010)—we only include the correspondences where the neighborhood is symmetric; that is, if $p$ and $q$ are each other's nearest neighbor in the local spectral feature space.

The spectral embeddings are $k$-dimensional representations of the original word embeddings that encode the local *knn* structure around each word. Since a linear transformation preserves the distances between all points, the spectral embeddings allow us to map each source word to a target word with a similar *knn* structure. The parameter $k$ offers a simple way to adjust the amount of contextual information used in building the spectral embeddings.

### 3.2 Estimating Global Correspondences

After extracting initial correspondences using spectral features, we iteratively update the mapping to preserve the global pair-wise distances using the iterative mapping (IM) algorithm. The objective of IM is to preserve the relative distances among the source words in the mapped space, which is achieved by locally minimizing a global loss function in iterations until convergence. Note that the spectral embeddings described in Section 3.1 are only used to extract tentative pairs for initialization. Since the spectral embeddings only capture local features, the rest of the algorithm uses the original word embeddings to preserve global distances among source words.

Given a set of $n$ monolingual embeddings $X$ for the source language, and a set of $m$ monolingual embeddings $Y$ for the target language, we use the residual sum of squares loss function defined below to optimize the mapping $M$ from $X$ to $Y$:

$$L = \sum_{p,q} \Big( D_X\big(x_p, x_q\big) - D_Y\big(M(x_p), M(x_q)\big) \Big)^2 \quad (2)$$

where $D_X$ and $D_Y$ are the pairwise Euclidean distances for $X$ and $Y$, respectively, and $p = 1, ..., n$, $q = 1, .., n$ span the indices in $X$.

We seed the learning using the correspondences obtained by the spectral initialization method. The remaining words are mapped to a virtual token with a distance $c$ from all other words, including itself, where $c > 0$ is a tunable parameter. The optimization is then carried out in a greedy manner: a source word, $x_i$, is selected at random, and $M(x_i)$ is selected to be the word in $Y$ that minimizes the loss function $L$. This greedy algorithm yields a locally optimal mapping at each step and the final result depends on the initialization. The IM method is summarized in Algorithm 1.

After optimizing the global distances using IM, we use the $(source, target)$ pairs in $M$ to learn a linear transformation between $X$ and $Y$ as described in Section 2.2.

**input** : Word embeddings X and Y
**output:** Mapping M from X to Y
$M \leftarrow spectral\_initialization(X, Y)$
$C \leftarrow cost\_of\_mapping(M, X, Y)$
**repeat**
    Sample a word $x \in X$
    **for** $y \in Y$ **do**
        $M_y \leftarrow M$
        $M_y(x) = y$
        $C_y \leftarrow cost\_of\_mapping(M_y, X, Y)$
        **if** $C_y < C$ **then**
            $M \leftarrow M_y$
            $C \leftarrow C_y$
        **end**
    **end**
**until** *convergence or max iterations*;
**Algorithm 1:** Iterative mapping with spectral initialization

## 4 Experiments

We experimented with two language pairs: French-English, and Arabic-English. French shares similar orthography and word roots with English, but for evaluating the generality of the approach, we don't utilize these similarities in any form.[1] Arabic, on the other hand, is a dissimilar language with more limited resources, and it is noisier at the word level due to clitic affixation that is challenging to tokenize. This makes it a suitable test-case for a realistic low-resource language.

### 4.1 Data

We extracted various datasets with different levels of similarity to test the proposed unsupervised word mapping approach. We used the following data sources:

**WMT'14** the Workshop on Machine Translation French-English corpus (Bojar et al., 2014). This is a parallel corpus, but we don't use the sentence alignments.

---

[1] The word embeddings are learned independently for each language; representations of subword units are not shared across languages, so morphological variations are only accounted for mono-lingually.

| Label | Source | Target |
|---|---|---|
| fr-en-p | French WMT'14 | English WMT'14 |
| fr-en-s | French AFP | English APW |
| fr-en-d | French APW 200x | English APW 199x |
| ar-en-p | Arabic UN | English UN |
| ar-en-s | Arabic AFP | English APW |

**Table 1:** French-English and Arabic-English datasets. *fr-en-p* and *ar-en-p* are parallel datasets, and the remaining are non-parallel. *fr-en-d* is extracted from separate time periods to ensure that there is no overlap in content.

**AFP** Agence France Presse corpora from Gigaword datasets for English (Parker et al., 2011b), French (Mendonça et al., 2009), and Arabic (Parker et al., 2011a).

**APW** The Associated Press corpora from Gigaword datasets.

**UN** Parallel Arabic-English corpus from UN proceedings (Ma, 2004)

We randomly extracted 5M sentences from each corpus to create the datasets in Table 1, which are either parallel (suffix:*p*), similar (suffix:*s*), or dissimilar (suffix:*d*). All datasets are within-genre to ensure that they share a common vocabulary. We tokenized the English and French datasets using the CoreNLP toolkit (Manning et al., 2014). We also converted all characters to lower case and normalized numeric sequences to a single token. Arabic text was tokenized using the Madamira toolkit (Pasha et al., 2014). We used the D3 tokenization scheme, and we further processed the data by separating punctuation and normalizing digits. Note that Arabic tokenization is non-deterministic due to clitic affixation, so the processed datasets still contained untokenized phrases.

### 4.2 Experimental Set-up

For each of the datasets described above, we generated 100-dimensional word embeddings using the subword skip-gram model (Bojanowski et al., 2016). We extracted the most frequent 2K words from the source and target languages and their embeddings for the iterative mapping (IM) method. The loss function $L$ in equation 2 was used to guide the tuning of model parameters. We tuned $k = [10, 20, 30, 40, 50]$ for the spectral initialization, and due to randomness in IM, we repeated each experiment 10 times and used the mapping that resulted in the smallest loss. For the final linear transformation $T$, we used the most frequent 50K words in both source and target languages, and we used the hubness reduction method described in (Dinu et al., 2015) with c=5000.

We extracted dictionary pairs from the Multilingual WordNet (Miller, 1995; Sagot and Fišer, 2008; Elkateb and Fellbaum, 2006; Abouenour et al., 2013) where the source words are within the top 15K words in all datasets. From these pairs, we extracted a random sample of 2K unique $(source, target)$ pairs for training the supervised method, and the remaining source words and all their translations were used for testing. This resulted in a total of 977 French words and 473 Arabic words for evaluation.

### 4.3 Analysis and Results

The unsupervised word mapping method proposed in this paper consists of three parts: given a subset of source and target words with a viable mapping, we extract tentative correspondences using spectral features as in Section 3.1. These initial pairs are used to seed the IM algorithm to refine the mapping as described in Section 3.2. The final correspondences obtained by the IM algorithm are then used as a seed dictionary to fit a linear transformation matrix between the source and target embeddings. The linear transformation step serves as a smooth generalization of the mapping since it preserves the structure of the source embeddings and can be used to extract translations of additional word pairs.

#### 4.3.1 Word Frequency Analysis

In order to extract a mapping between two sets of points, we first need to ensure that a viable mapping between the two sets exists. In an unsupervised setting, we can analyze the word frequencies within the monolingual corpora; it is reasonable to assume that certain words would have high frequencies in multilingual datasets that cover similar topics. Word frequencies follow a consistent power distribution that is at least partially determined by meaning (Piantadosi, 2014). Using a set of 200 fundamental words, (Calude and Pagel, 2011) reported a high correlation between word frequency ranks across 17 languages drawn from six language families.

We analyzed the consistency of word frequencies in the French-English dataset *fr-en-s* using all Word-

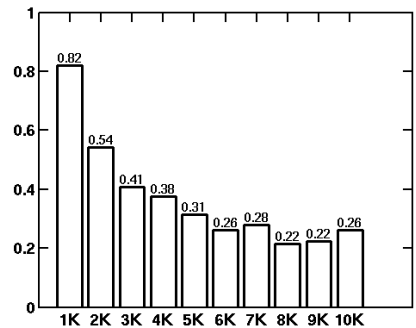

**Figure 1:** Frequency overlap (percentage of WordNet source words that have a translation within the same frequency band) in *fr-en-s* dataset.

Net translation pairs where source words fall within certain frequency bands. For example, given all French words in WordNet that fall within the 1K most frequent words, we report the fraction of these words that have a translation within the 1K most frequent words in English. Among the top 10K source words, we have a total of 4,653 words with WordNet translations, almost equally distributed among the ten frequency bands.

As show in Figure 1, at least 80% of the most frequent 1K French words have a translation within the same frequency band. Smaller overlap is observed for lower frequencies, where only about a quarter of the words have a translation within the same frequency band. This both confirms previous findings about the correlation of frequency ranks across different languages and also indicates that the correlation itself is dependent on word frequency. Note also that frequency ranks for the least frequent words are rather meaningless since most words in any finite dataset are likely to occur only once. Therefore, we carry our analysis and mapping using only the top 2K source and target words to improve the chances of having a feasible mapping between the two point sets.

#### 4.3.2 Nearest Neighbor Structures

To extract initial correspondences, we assume that similar words have similar *knn* graphs. Figure 2 shows colormap visualizations of *knn* adjacency matrices of various source and target words in *fr-en-s*, where red represents higher similarity scores close to 1.

| Source | Translation | Initial mapping |
|---|---|---|
| organisation | organization | development |
| agence | agency | office |
| dit | say | with |
| chine | china | singapore |
| project | plan | questions |
| ouest | west | amid |
| victimes | victims | death |
| partir | go | asked |
| refusé | refused | named |
| conflit | conflict | elections |
| constitution | constitution | democracy |

**Table 2:** A sample of initial pairs extracted using spectral embeddings to initialize IM for *fr-en-s*. **Source** indicates the source French word, **Translation** is the gold English correspondent, **Initial Mapping** is the first locally induced correspondent.

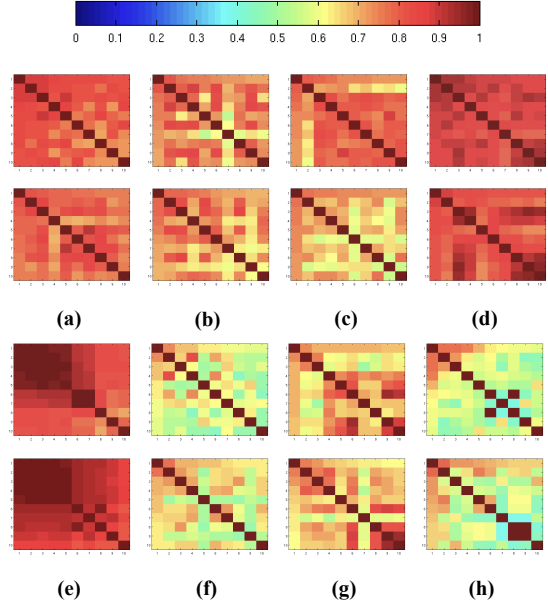

**Figure 2:** colormaps of knn adjacency matrices (k=10) of corresponding English (top) and French (bottom) words: (a) "go" - "partir" (b) "refused" - "refusé" (c) "hold" - "tenir" (d) "say" - "dit" (e) "monday" - "lundi" (f) "office" - "agence" (g) "china" - "chine" (h) "university" - "université".

Most words have similar color distributions in their neighborhood graphs as their translations, although most of them are not sufficiently distinct from other words, which is expected given the small dimensionality of the spectral space. Note also that most verbs have dense adjacency graphs due to variations in conjugation that tend to be clustered densely in the vector space. Ambiguous verbs like *hold* have dissimilar local structures, which reflects their inconsistent usage across the two languages. Nouns, on the other hand, tend to have less dense and more distinct local structures. One exception here is *monday* whose closest neighbors are other days of the week that have very similar representations, which results in a dense but consistent structure.

Figure 3 shows two-dimensional projections of original word embeddings and their corresponding spectral embeddings. Note that most words moved closer to their correct translations in the spectral space, where words with similar adjacency graphs are clustered in the same regions.

Table 2 shows a sample of initial correspondences extracted using spectral features for IM initialization. As expected, most word pairs are incorrectly mapped but semantically related to the target translation.

### 4.3.3 Global Distances

To verify the consistency of global distances, we randomly extracted a set of 100 WordNet pairs that lie within the most frequent 2K words in *fr-en-s*, and we divided the set into two sets of 50 words each and calculated the pair-wise Euclidean distances among the English words (Figure 4a) and among the corresponding French words (Figure 4b). For comparison, we extracted an additional random set of French words and calculated the Euclidean distances among them (Figure 4c). As shown, the colormaps of corresponding English and French words are relatively similar compared to random words, which indicates that global pairwise distances also reflect consistent language-independent features.

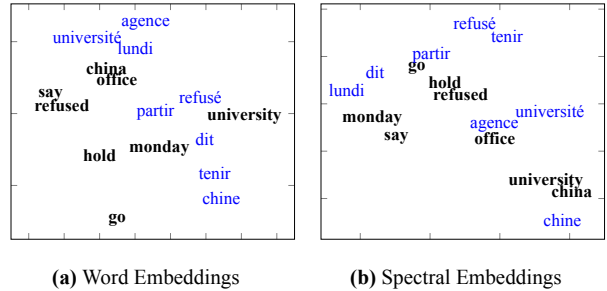

**Figure 3:** PCA projections of (a) word embeddings and (b) spectral embeddings of English (black, boldface) and French (blue) words from *fr-en-s* dataset.

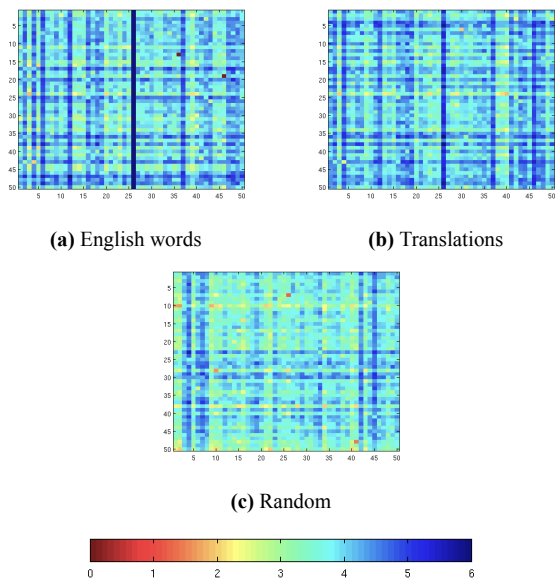

**Figure 4:** Colormaps of Euclidean distances between random sets of (a) English words, (b) their corresponding French translations, and (c) French words that are not translations of the words in (a). The Euclidean distance matrices shown here are asymmetric; the horizontal and vertical directions correspond to disjoint sets of 50 words each within the same language, for a total of 50 × 50 distances.

| Source | Translation | IM | | |
|---|---|---|---|---|
| | | fr-en-p | fr-en-s | fr-en-d |
| procureur | prosecutor | judge | *prosecutor* | attorney |
| difficultés | difficulties | differences | *problems* | conditions |
| février | february | *february* | september | january |
| rue | street | *street* | scene | square |
| véhicule | vehicle | port | bus | bus |
| demander | ask | *ask* | *ask* | *ask* |
| avenir | future | challenge | opportunity | ways |
| locale | local | *local* | *local* | central |
| crime | crime | trafficking | *criminal* | murder |
| continue | continuous | *continues* | comes | seeking |
| partis | parties | power | *parties* | *groups* |
| cinq | five | seven | three | three |
| mère | mother | woman | wife | *mother* |
| permet | allow | *allows* | used | used |

**Table 3:** A random sample of word mappings from French to English using IM with spectral initialization. These pairs are later used to fit a linear projection matrix between the source and target embeddings. *Correct mappings are indicated in italics*

| Source | Translation | IM | |
|---|---|---|---|
| | | ar-en-p | ar-en-s |
| عليا | supreme | *high* | *supreme* |
| دبلوماسي | diplomatic | peaceful | justice |
| توسيع | expand | *expansion* | *boost* |
| منافسة | competition | sound | player |
| أعربت | expressed | *expresses* | comment |
| طلبت | asked | *requests* | *demanded* |
| معلومات | information | *information* | *information* |
| فرصة | chance | *opportunity* | *chance* |
| تدريب | training | centres | *training* |
| شمال | north | west | southern |

**Table 4:** A random sample of word mappings from Arabic to English retrieved using IM with spectral initialization. *Correct mappings are indicated in italics*

Tables 3 and 4 show a subset of word mappings retrieved using IM with spectral initialization on the various datasets. Recall that the objective of IM is to preserve global pairwise distances of the source words in the mapped space. Most IM mappings are either correct or related to the target translation; for example, the French word for February is mapped to September or January, which are nearest neighbors of the correct word in the target vector space and are semantically related. Using samples of 100 words randomly extracted from each dataset, we estimated the quality of word translations in terms of semantic similarity and relatedness. [2] As seen in Figure 5, over 60% of translations are semantically related, of which at least 20% are semantically similar.

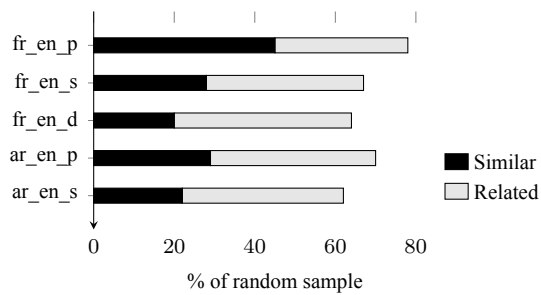

**Figure 5:** Quality estimation of IM-SI word translations.

---

[2] words are considered semantically similar if they are synonymous or identical in meaning regardless of syntactic category; for example {*happy, glad, happiness*}. Semantically related words are somewhat related in meaning but not necessarily synonymous, such as {*food, fruit, restaurant*}.

### 4.3.4 Linear Transformation

Learning optimal linear transformations between multilingual vector spaces depends on the quality and size of the seed dictionaries while unsupervised mappings are expected to be noisy. In this section, we evaluate the quality of linear transformations with suboptimal supervision. Figure 6 demonstrates the performance of the transformations learned using dictionary pairs extracted from wordNet with different sizes and perturbation levels. The performance is reported in terms of precision at $k$, where $k$ is the number of nearest neighbors in the target vocabulary.

Larger dictionaries result in more accurate transformations as expected. A thousand or more accurate dictionary pairs are sufficient to learn high quality transformations, while smaller dictionary sizes result in much lower precision at all $k$ levels. Figure 6b shows the performance using a training dictionary of size 2K perturbed with incorrect mappings. Surprisingly, the precision is reasonably high even when only 50% of the dictionary pairs are correct. This indicates that a bilingual transformation can be learned successfully using few thousand word pairs even in the presence of noise, so a reasonable amount of incorrect mappings can be tolerated.

## 5 Evaluation

Using the $(source, target)$ pairs extracted using IM with spectral initialization (IM-SI), we fit a linear projection matrix from the source to the target embeddings to compare the results with supervised linear transformation. We also compare with a baseline of random initialization of the IM method (IM-Rand). We evaluate the linear transformations on the different datasets in Table 1 by reporting the precision of mapping each test word to a correct translation within its $k$ nearest neighbors, for $k \in \{1, 5, 10, 20, 50, 100\}$. The results are shown in Figure 7.

While the initial spectral embeddings didn't always recover the correct correspondences (see Table 2), these tentative pairs helped initialize the IM algorithm in the right direction for better global convergence. As shown in Figure 7, initializing IM with random pairs resulted in poor performance while spectral initialization helped converge to plausible

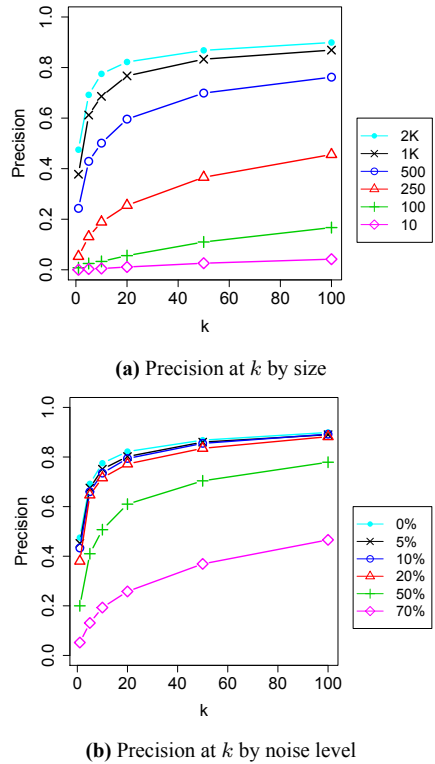

**(a)** Precision at $k$ by size

**(b)** Precision at $k$ by noise level

**Figure 6:** Bilingual transformation precision at $k$ with different characteristics (size and noise level) of the seed dictionary. The transformations are learned on *en-fr-s* word embeddings

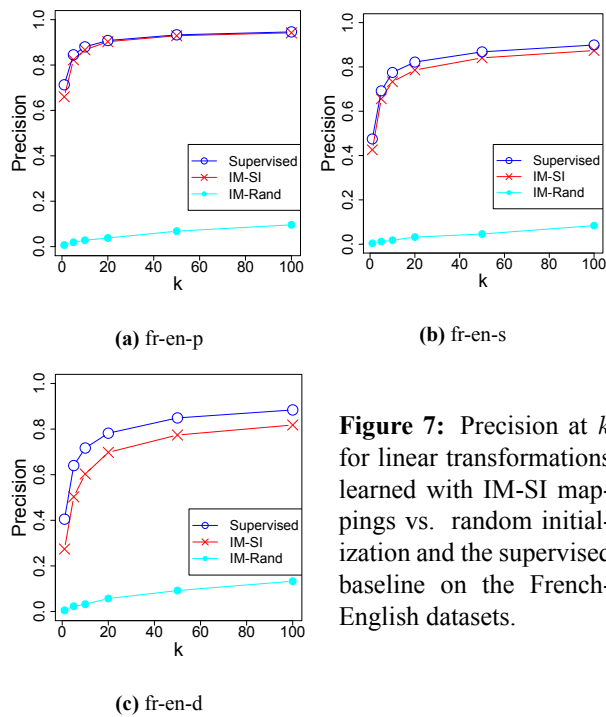

**(a)** fr-en-p

**(b)** fr-en-s

**(c)** fr-en-d

**Figure 7:** Precision at $k$ for linear transformations learned with IM-SI mappings vs. random initialization and the supervised baseline on the French-English datasets.

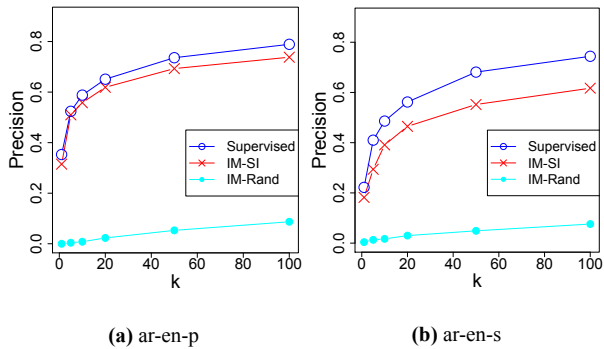

**(a)** ar-en-p      **(b)** ar-en-s

**Figure 8:** Precision at $k$ for linear transformations learned using IM-SI vs. IM-Rand and the supervised baseline on the parallel and non-parallel Arabic-English datasets

| Source | WordNet Targets | *knn*s of transformed vector | |
|---|---|---|---|
| | | **ar-en-p** | **ar-en-s** |
| – Correct – | | | |
| حوار | 'negotiation', 'argumentation', 'dialogue', 'argument', 'talks', 'debate' | 'dialogue', 'dialogues', 'consensus-building', 'intra-east', 'all-inclusive' | 'dialogue', 'dialogues', 'peace', 'conciliation', 'reconciliation' |
| مجلة | 'journal', 'magazine' | 'co-author', 'publishes', 'journal', 'magazine', 'publisher' | 'magazine', 'publishes', 'edition', 'newsweek', 'publisher' |
| – Incorrect – | | | |
| مطلب | 'claim', 'requirement', 'prerequisite', 'demand' | 'principled', 'onus', 'insists', 'insisting', 'rests' | 'insistence', 'objection', 'demands', 'refusal', 'deference' |
| بناية | 'building', 'edifice' | '<num>-bed', 'floors', 'tower', 'dormitory', 'playground' | 'parking', 'three-story', 'six-story', 'mall', 'five-story' |

**Table 5:** Examples of correct and incorrect transformations at $k = 5$ for Arabic-English using the unsupervised IM-SI mappings to fit a linear projection matrix.

mappings. In fact, the use of spectral initialization in combination with IM to seed the transformation resulted in a precision close to the supervised baseline as seen in Figures 7a and 7b.

Figure 8 shows the performance of transforming Arabic word embeddings using the various models. The supervised baseline results are lower than the French-English case, which is partly due to the low coverage of WordNet translations for Arabic (see Table 5). Nevertheless, we managed to recover accurate mappings and linear transformations that perform comparably to the supervised baseline. Table 5 shows some examples of correct and incorrect transformations at $k = 5$ on Arabic test words. Observe that even in the case of incorrect matches, the $k$ nearest neighbors are related to the target words in meaning. For example, all five nearest neighbors of the word ('بناية'/'building'), are building-related, such as 'tower', 'parking', 'three-story', and 'mall'.

## 6 Conclusion

We proposed an unsupervised approach for learning linear transformations between word embeddings of different languages without the use of seed dictionaries or any prior bilingual alignment. The proposed method exploits various features and structures in monolingual vector spaces, namely word frequencies, local neighborhood structures, and global pairwise distances, assuming that these structures are sufficiently consistent across languages. We verified experimentally that, given comparable multilingual corpora, accurate transformations across languages can be retrieved using only their monolingual word embeddings for clues.